\documentclass[runningheads]{llncs}

\usepackage[T1]{fontenc}

\usepackage{graphicx}
\usepackage{subfigure}

\usepackage{accv}
\usepackage{accvabbrv}
\usepackage{multirow}
\usepackage{booktabs}
\usepackage{latexsym}

\usepackage[breaklinks,colorlinks,citecolor=green]{hyperref}
\usepackage{color}

\usepackage{enumerate}
\usepackage{paralist}

\begin{document}

\title{UNet$--$: Memory-Efficient and Feature-Enhanced Network Architecture based on U-Net with Reduced Skip-Connections}

\titlerunning{UNet$--$:U-Net with Reduced Skip-Connections}

\author{Lingxiao Yin\thanks{Corresponding author.}\inst{1} \and Wei Tao\inst{1} \and Dongyue Zhao\inst{1} \and Tadayuki Ito\inst{2} \and Kinya Osa\inst{2} \and Masami Kato\inst{2} \and Tse-Wei Chen\inst{2}
}

\authorrunning{Yin, L., et al.}

\institute{Canon Innovative Solution (Beijing) Co., Ltd.\\ 
\email{\{Lingxiao\_Yin, taowei, zhaodongyue\}@canon-is.com.cn} \and Device Technology Development Headquarters, Canon Inc. \\
\email{twchen@ieee.org}
}

\maketitle

\begin{abstract}
U-Net models with encoder, decoder, and skip-connections components have demonstrated effectiveness in a variety of vision tasks. The skip-connections transmit fine-grained information from the encoder to the decoder. It is necessary to maintain the feature maps used by the skip-connections in memory before the decoding stage. Therefore, they are not friendly to devices with limited resource. In this paper, we propose a universal method and architecture to reduce the memory consumption and meanwhile generate enhanced feature maps to improve network performance. To this end, we design a simple but effective Multi-Scale Information Aggregation Module (MSIAM) in the encoder and an Information Enhancement Module (IEM) in the decoder. The MSIAM aggregates multi-scale feature maps into single-scale with less memory. After that, the aggregated feature maps can be expanded and enhanced to multi-scale feature maps by the IEM. By applying the proposed method on NAFNet, a SOTA model in the field of image restoration, we design a memory-efficient and feature-enhanced network architecture, UNet$--$. The memory demand by the skip-connections in the UNet$--$ is reduced by 93.3\%, while the performance is improved compared to NAFNet. Furthermore, we show that our proposed method can be generalized to multiple visual tasks, with consistent improvements in both memory consumption and network accuracy compared to the existing efficient architectures.

\keywords{Skip-connection \and Memory \and U-Net.}
\end{abstract}

\section{Introduction}
\label{sec:introduction}
\begin{figure}[htbp]
	\centering
	\subfigure[]{
		\label{fig:U-Net}
		\includegraphics[width=0.45\textwidth]{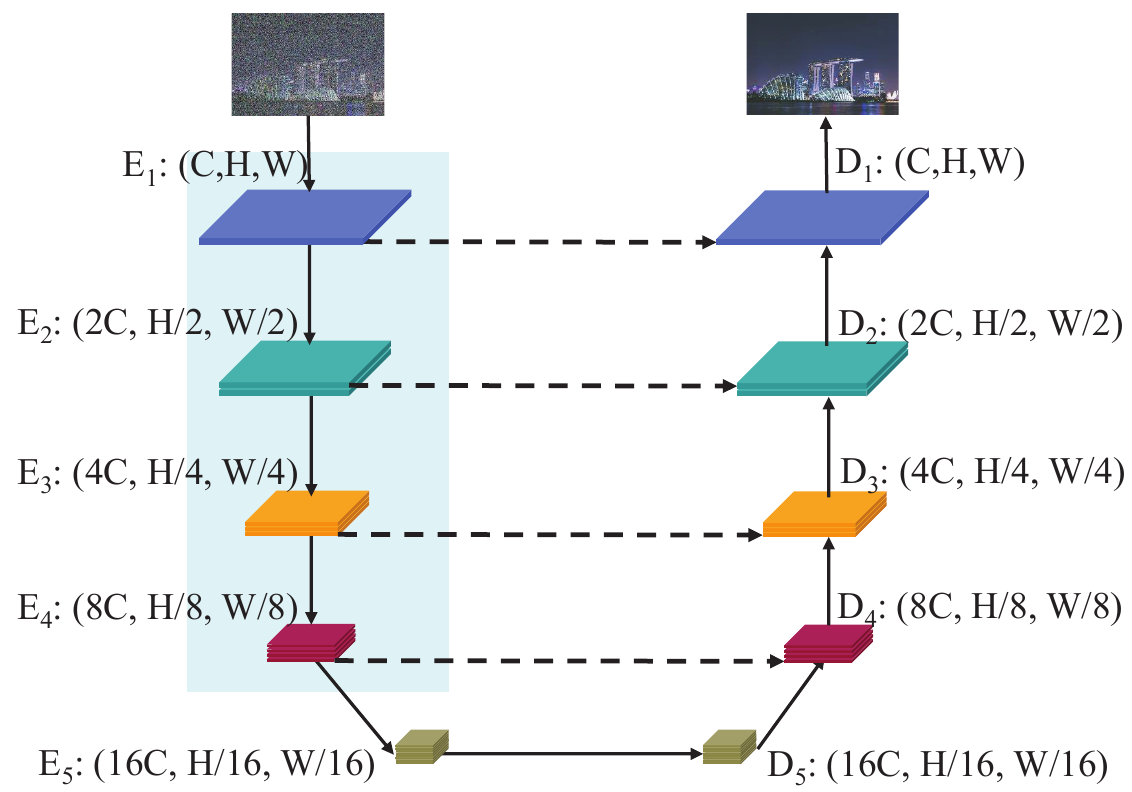}}
	\quad
	\subfigure[]{
		\label{fig:Memory}
		\includegraphics[width=0.45\textwidth]{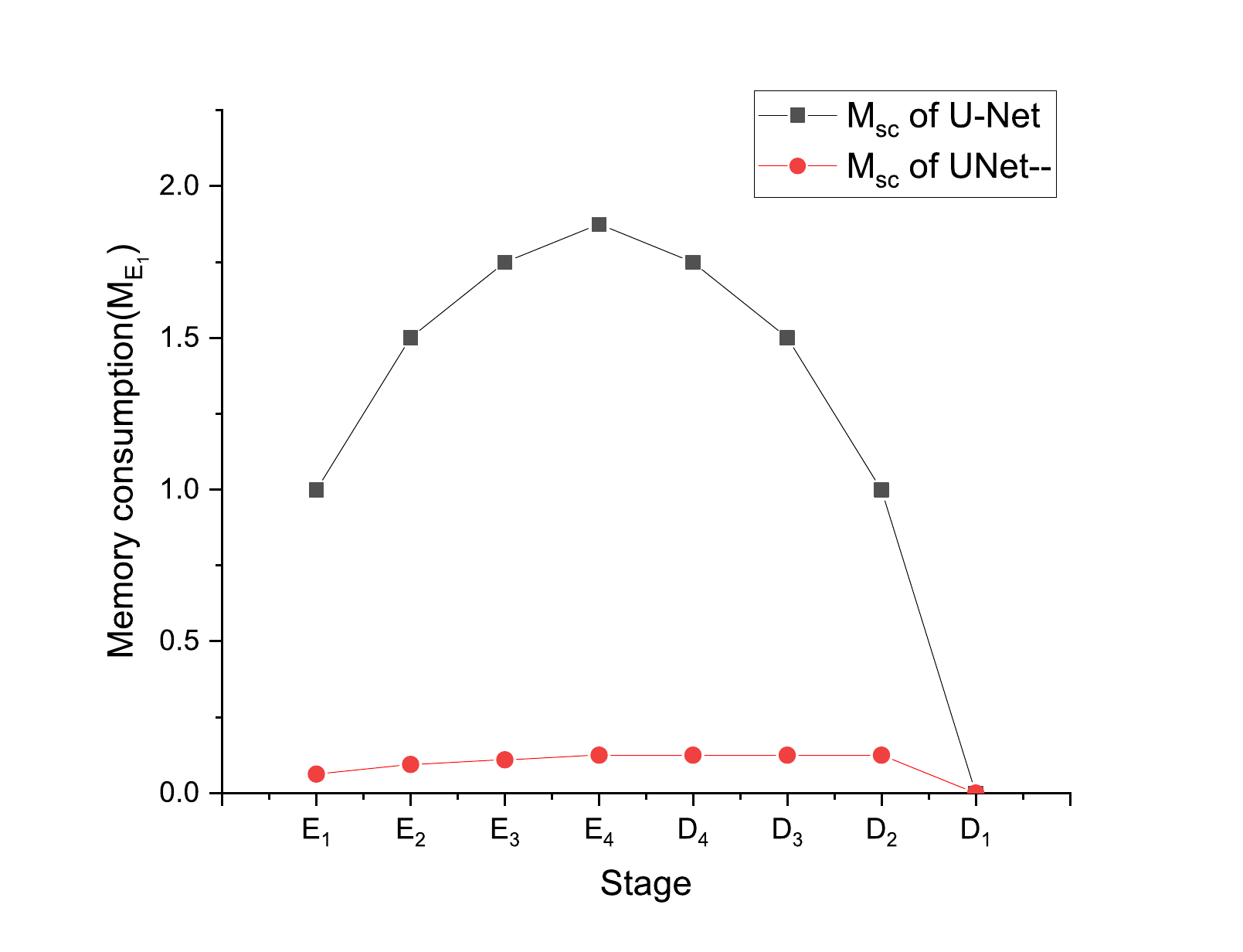}}
	\caption{(a) Architecture of a common U-Net model. The dashed lines represent the skip-connections. Feature maps in the blue shadow are needed to be maintained in memory. (b) Memory consumption for skip-connections for U-Net and UNet$--$ throughout the inference process, the unit is $M_{E_1}$ which denotes the memory consumption of $E_1$. }
	\label{fig:U-NetAndMemory}
\end{figure}

U-Net\cite{2015U} and its variants\cite{unet++,UNet3+,iek20163D,oktay2018attention,2016Rethinking} have become the de facto standard in the field of pixel-level computer vision research\cite{Zhou_2021}. With the ability to learn effective features efficiently and robustly, they achieve superior performance in many tasks \eg, image restoration\cite{chen2022simple}, image segmentation\cite{Guo2022SegNeXtRC,YOLACT,SparseInstance}, image matting\cite{Mask_Mask_matting}, \etc. U-Net consists of an encoder, a decoder, and skip-connections between them as in \cref{fig:U-Net}. Feature maps are downsampled stage by stage in the encoding phase and correspondingly upsampled in the decoding phase. The skip-connections transmit the details of the encoding information to the decoder. They can compensate for details loss due to downsampling operations with a tolerable computation cost. However, they are not friendly in terms of memory consumption in especially resource-limited device. The feature maps transmitted by the skip-connections require on-chip memory and the memory will not be released until the feature maps of the skip-connections are fused into the decoder. Therefore, despite their great success, deploying U-Net or its variants on resource-limited devices is still a non-trivial task.

To reduce the heavy memory consumption induced by the skip-connections, recently many researchers have proposed methods of efficient model design to attempt to merge them into the parallel convolutional structures. RepVGG\cite{ding2021repvgg} proposes structural re-parameterization to decouple a training-time multi-branch topology with an inference-time plain architecture. Inspired by the method, the authors in \cite{FMEN,mobileone2022,vasufastvit2023} merge the skip-connections into a parallel convolutional structure during inference. The final obtained single-branch structure is equivalent to the original multi-branch structure, as there are no non-linear units in the original structure. Although they reduce the memory consumption and improve inference speed, the methods are not appropriate for the skip-connections in U-Net yet. The main reason is that in general in the multi-branch structure represented by the skip-connections, one of the branches on the main trunk contains several non-linear transformations. Consequently, it is not achievable to merge the skip-connections equally. 

The authors in \cite{Tailor, Singh2024} propose to reconstruct a model without skip-connections which is obtained by progressively removing skip-connections during finetuning. The model is trained with the distillation method to mitigate performance drop, where the teacher model is combined with skip-connections. Here the skip-connections often refer to the shortcuts in residual structure, where
the distance between layers connected by skip connections is approximately two or three layers. In contrast, the layers which are connected by the skip-connections in U-Net are very far away. It spans across two stages of the encoder and the decoder, and it often leads to a large difference in the feature maps at both endpoints of the skip-connections. Directly removing such skip-connections from U-Net can result in significant accuracy loss, which is verified in  \cref{tab:denoising,tab:deblurring,tab:img_resolution,tab:img_matting}. It is important to reduce the memory consumption of the skip-connections without introducing harm in accuracy. 

Some other works aim to improve U-Net performance by designing complex structures\cite{Deformable4, Deformable-residual-image-super-resolution, zhang2020resnest, ibtehaz2023accunet} to replace the simple skip-connections. The authors in \cite{ibtehaz2023accunet} found that the combination of feature maps from multiple encoder levels generates enriched features compared to the simple skip-connection. This benefits from the combination of various regions of interest from different levels. However, these complex structures lead to high computation and memory usage.	
We propose an efficient multi-scale information aggregation module (MSIAM) and an effective information enhancement module (IEM) which can be easily integrated into any U-Net or its variants. Specifically, the MSIAM first reduces the number of channels in multi-scale feature maps to save memory, then these compressed feature maps are resized to a single-scale with the same resolution and concatenated together, and finally information interaction is achieved through point-wise convolution.  The generated single-scale feature map is maintained in memory instead of the original large feature maps. The IEM moduel is designed to generate enhanced multi-scale feature maps based on the single-scale feature map. First, the single-scale feature map is resized to multiple groups of feature maps with different resolutions, and then an enhancement block is applied to each group to generate enriched feature maps. To avoid a large computational cost, we introduce a ConvNeXt V2 block and a separable convolution in the enhancement block.

By applying MSIAM and IEM to U-Net, we construct a memory-efficient and feature-enhanced network architecture named UNet$--$.

The main contributions are summarized as follows:
\begin{compactitem}
	\item[1.] We significantly reduce the memory consumption induced by the skip-connections in U-Net and its variants, and further construct a memory-efficient and feature-enhanced network architecture.
	\item[2.] We design plug-and-play modules, MSIAM and IEM, which are applicable to any U-Net and its variants. 
	\item[3.] We demonstrate the proposed method in a great variety of tasks and verify its generality and effectiveness.
\end{compactitem}

\section{Related work}

The skip-connection, also known as shortcut or residual, is a popular trick in neural network design. It was put forward by ResNet\cite{ResNet} in which the layer is reformulated as residual learning function with reference to the input of the layer by introducing an identity mapping. The design of skip-connections makes training deeper networks easier and gains accuracy from the increased depth. Inspired by this idea, the skip-connection is widely used in latter networks\cite{Inception-v4}. Inception-v4\cite{Inception-v4} significantly accelerates Inception network training by skip-connections. ResNeXt\cite{ResNeXt} combines skip-connections with designed homogeneous multi-branch architecture. Res2Net\cite{Res2Net} constructs hierarchical residual-like connections within one single residual block. 

Based on FCN\cite{long2015fully}, U-Net\cite{2015U} introduces long-path skip-connections between the contracting path and the expansive path. Such network design obtains great success in pixel-level prediction task (e.g. semantic segmentation, instance segmentation, image restoration) as the skip-connections can transmit fine-grained information lost caused by downsampling. In order to obtain higher accuracy, some designs build more complex connections. \cite{unet++, UNet3+}, introduce attention mechanism for the feature maps of skip-connections\cite{oktay2018attention, Attention_Unet++, sun2023datransunet}. Some recent work designs hybrid block to model global and multi-scale context\cite{UCTransNet},\cite{ibtehaz2023accunet}. 

Although we can benefit from skip-connections, large memory consumption is required. These works\cite{ding2021repvgg, mobileone2022, vasufastvit2023} demonstrate that through structural re-parameterization, skip-connections can be merged into their parallel convolutions to reduce extra memory requirements. \cite{ding2021repvgg} is a simple architecture with a stack of 3 × 3 conv and ReLU. During the training process, it stacks a sets of multi-branch structures composed of 3 × 3 conv, 1 × 1 conv, and identity mapping. After training, it converts the trained multi-branch structure into a single 3 × 3 conv for inference. Consequently, the converted model only has a stack of 3 × 3 conv, which is efficient for test and deployment. \cite{mobileone2022} is another technique that converts multi-branch networks with skip-connections to plain networks through structural re-parameterization and over-parameterization. \cite{vasufastvit2023} uses structural re-parameterization to lower the memory access cost by removing skip-connections in the network based on transformer architecture. Some other works are proposed to progressively remove skip-connections during training process \cite{Tailor} or further with distillation\cite{Singh2024}. \cite{Tailor} introduces two new methods, SkipRemover and SkipShortener, that alters networks with skip-connections dynamically during retraining by remove or shorten the skip connections to fit better on hardware, achieving resource-efficient inference with minimal to no loss in accuracy. As analyzed in \cref{sec:introduction}, these methods are not applicable to skip-connections with long path in U-Net networks. In this paper, we propose a different method to significantly decrease memory consumption caused by skip-connections, a set of multi-scale feature maps are transformed into a lightweight single-scale one, which will be expanded and enhanced later.

\section{Methodology}
In this chapter, we introduce the runtime memory analysis of common U-Net models in  \cref{sec:Memory_analysis}. We present the proposed UNet$--$ architecture and its details in  \cref{sec:OverviewModel} and \cref{sec:architecture_design}.

\subsection{Memory analysis}
\label{sec:Memory_analysis}
The common U-Net model and its runtime memory analysis is shown in \cref{fig:U-NetAndMemory}. There are five stages in the encoder and decoder, respectively. The feature maps output at each stage in the encoding process are marked as  \{$E_1, E_2, E_3, E_4, E_5$\} . Between adjacent stages, by downsampling operations the feature map size is halved and the number of channels is doubled. Similarly, the feature maps output at each stage in the decoding process are marked as  \{$D_1, D_2, D_3, D_4, D_5$\}. Between adjacent stages, through upsampling operations, the feature map size is doubled and the number of channels is halved. There are four skip-connections that transmit the feature maps from $E_1$ to $D_1$,  $E_2$ to $D_2$,  $E_3$ to $D_3$, and $E_4$ to $D_4$, respectively. We note the total memory size for feature maps to be transmitted via skip-connection as $M_{sc}$. 
To simplify the analysis, we ignore the memory required for loading model parameters as its value is constant. We consider the value of $M_{sc}$ at the end of each stage of the encoder and decoder processes. At the end of $E_1$, the output feature maps need to be maintained in memory for later connection to $D_1$. The size of the feature maps is calculated as \cref{eq: E1_memory_size}. 
\begin{equation}
	\label{eq: E1_memory_size}
	M_{E_1} = C\times H \times W 
\end{equation}
then $M_{sc}=M_{E_1}$ at the end of $E_1$. Similarly at the end of $E_2$, the output feature maps also need to be maintained for later connection to $D_2$ whose memory size is:
\begin{equation}
	\label{eq: E2_memory_size}
	M_{E_2} = 2\times C \times \frac{H}{2}\times\frac{W}{2} =\frac {M_{E_1}}{2}
\end{equation}
then $M_{sc}=M_{E_1} + M_{E_2}=\frac {3 \times M_{E_1}}{2}$. Analogously, $M_{sc}=M_{E_1} + M_{E_2} +  M_{E_3}=\frac {7 \times M_{E_1}}{4}$ at the end of $E_3$, and $M_{sc}=M_{E_1} + M_{E_2} + M_{E_3} +  M_{E_4}=\frac {15 \times M_{E_1}}{8}$ at the end of $E_4$. The memory of maintaining feature maps keeps increasing in the encoder process. It reaches a peak value of $ \frac{15}{8}\times M_{E_1}$ at the end of $E_4$.
At the end of $D_4$, the maintained feature maps from $E_4$ is released. Thus, at that time $M_{sc}$ starts to be decreased to $M_{E_1} + M_{E_2}+ M_{E_3}$. Along with the entire decoding process, $M_{sc}$ continues to be decreased to zero until the end the $D_1$. 
We illustrate the temporal distributive characteristics of $M_{sc}$ in \cref{fig:Memory}. It is worth to note that the $M_{E_4}$ is only $M_{E_1}$$\times \frac{1}{8}$. That means the skip-connections represented by $M_{sc}$ consume extra $14$ times memory of $M_{E_4}$ when it reaches peak value. Most of the memory will not be released until the decoding process is finished. In conclusion, the skip-connections present a great challenge for model deployment in resource-limited devices. 

\subsection{Overview of the method}
\label{sec:OverviewModel}
\begin{figure}[t]
	\centering
	\includegraphics[width=0.9\textwidth]{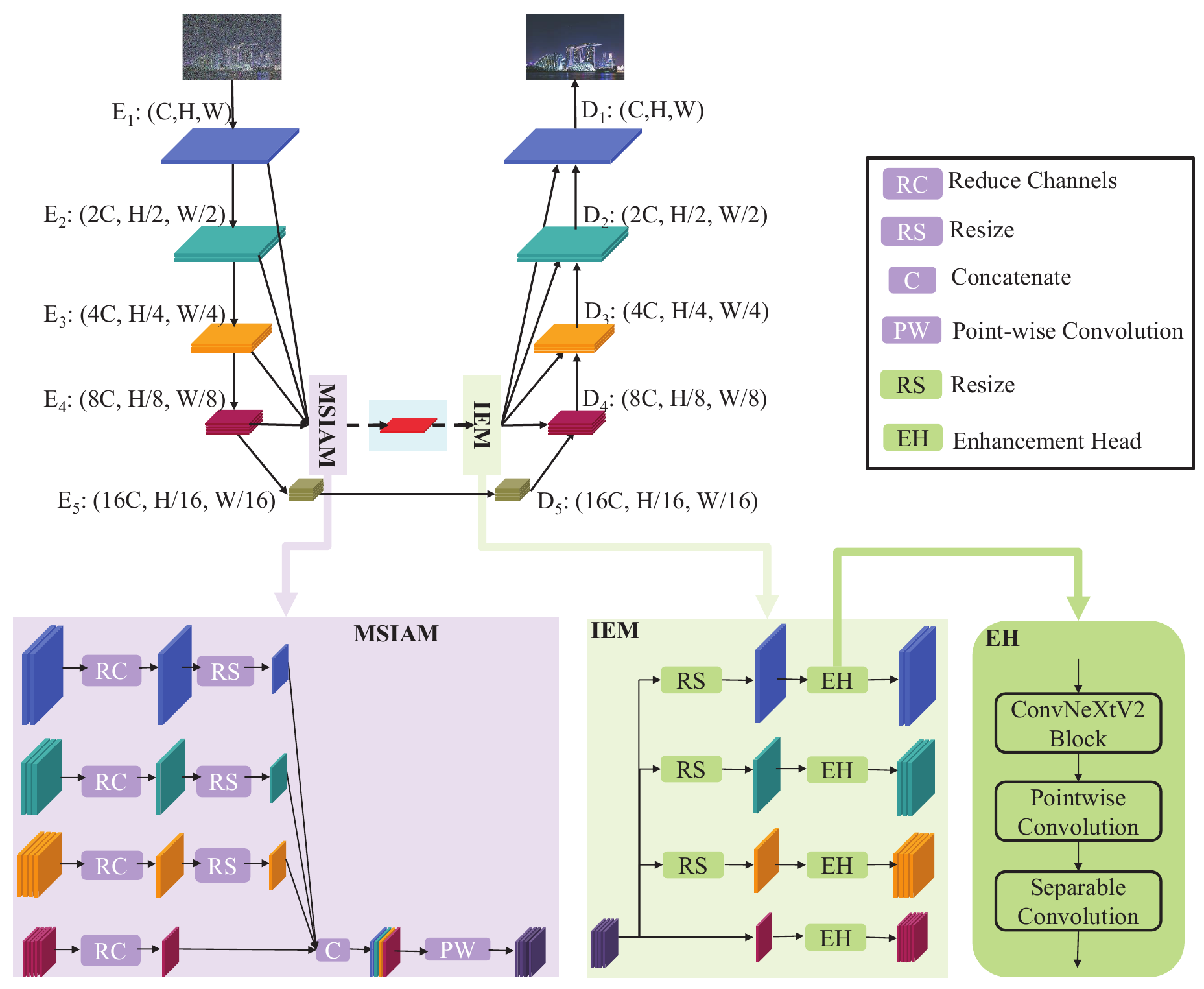}
	\caption{Illustration of the proposed UNet$--$ network. The original skip-connections are replaced with MSIAM and IEM. MSIAM aggregates multi-scale feature maps to single-scale with less memory demand. IEM generates enhanced multi-scale feature maps according to the output of MSIAM. Feature maps in the blue shadow are needed to be maintained in memory.}
	\label{fig:overview}
\end{figure}

We propose a universal method to reduce memory consumption caused by the skip-connections and meanwhile generate enhanced feature maps which boost network performance. Applied with our method, we design a memory-efficient and feature-enhanced network architecture based on U-Net, UNet$--$, which is named for consuming less memory by skip-connections than U-Net.
The model architecture is shown in \cref{fig:overview}. 

There are two main modules in UNet$--$: Multi-Scale Information Aggregation Module (MSIAM) and Information Enhancement Module (IEM). Unlike U-Net models, in UNet$--$ the multi-scale feature maps output from the encoder are not directly transmitted to the decoder. They are fed into the proposed MSIAM and become lighter sequentially. MSIAM aggregates the light-weight multi-scale feature maps and generates the corresponding single-scale feature maps but with information from multiple receptive fields. The single-scale feature maps require less memory than common skip-connections. During the decoding process, according to the input single-scale feature maps, IEM can generate a group of multi-scale feature maps. The generated multi-scale feature maps carry enhanced information with stronger representative ability. Through this way, the memory demand for skip-connections is significantly reduced to single-scale feature maps and the enhanced information generated by IEM helps to improve the model performance.
The macro architecture of the proposed modules is described in the following section.

\subsection{Macro architecture design}
\label{sec:architecture_design}
\subsubsection{Multi-Scale Information Aggregation Module (MSIAM)}

The target of this module is to generate a group of single-scale, compact, and representative feature maps. The memory footprint of the feature maps should be much smaller than that of common skip-connections.
\setlength{\tabcolsep}{10pt}
\begin{table}[t]

	\centering
	\begin{tabular}{@{}lll@{}}				
		\toprule
		Models & PSNR & SSIM \\
		\midrule
		Model-0 (w/o skip-connection)& 39.4737& 0.9561\\
		Model-1 ($E_1 \rightarrow D_1$)& 39.7906& 0.9587\\
		Model-2 ($E_2 \rightarrow D_2$)& 39.8738& 0.9593\\
		Model-3 ($E_3 \rightarrow D_3$)& 39.8703& 0.9593\\
		Model-4 ($E_4 \rightarrow D_4$)& 39.8111& 0.9587\\
		\bottomrule
	\end{tabular}
	\caption{Performance comparison with only one skip-connection, where ($E_i \rightarrow D_i$) represents the skip-connection connects $E_i$ and $D_i$. The results are evaluated on the SIDD validation dataset. 
	}\label{tab:single_sk_contribution}

\end{table}

To reduce the memory consumption caused by skip-connections, we are considering whether each skip-connections is crucial and whether it is possible to remove certain skip-connections directly. We first study the contribution of each skip-connections. We construct four U-Net models; each model has only one skip-connection between encoder and decoder. The four U-Net models share the same architecture except the position of the single skip-connection. Specifically, the single skip-connection in Model-1 connects $E_1$ and $D_1$, the skip-connection in Model-2 connects $E_2$ and $D_2$, the skip-connection in Model-3 connects $E_3$ and $D_3$, and the skip-connection in Model-4 connects $E_4$ and $D_4$. Besides, we also construct Model-0 without any skip-connections for comparison. We verify the performance of these five models in the image denoising task. The results of the experiment are shown in \cref{tab:single_sk_contribution}. We find that the contribution ratio is different and irregular. This is similar to the findings in \cite{UCTransNet} where they found that the optimal combination of skip-connections is different for different datasets. To obtain a universal structure which is appropriate with multiple tasks and datasets, it is useless to find an optimal combination of skip-connections. In contrast, we should keep all skip-connections to avoid the performance dropping dramatically.

In addition, the skip-connections in U-Net are independent of each other, which leads to the information with different receptive fields lacking interactivity. Referring to the ideas in \cite{unet++,UNet3+, UCTransNet,2017Feature,ibtehaz2023accunet}, we think that it is potential to generate powerful and robust information by combining all these multi-scale feature maps and directing the information to cooperate with each other automatically.

Taking these considerations into account, we design MSIAM in \cref{fig:overview}. There are four units in MSIAM that can be represented as \cref{eq:aggregation}.
\begin{equation}
	\label{eq:aggregation}
	E'=  {\rm PWConv}({\rm RS}({\rm RC}(E_1))||{\rm RS}({\rm RC}(E_2))||...{\rm RS}({\rm RC}(E_n))...||{\rm RS}({\rm RC}(E_N)))
\end{equation}
where, $E'$ is single-scale output feature maps of MSIAM, N and n represents how many stages in encoder and the current stage index, respectively, $E_n$ represents the feature maps outputted by the nth stage, PWConv is a point-wise convolution, RS means the resizing operation which adjusts the resolution of feature maps, RC means reducing channels of feature maps, $||$ represents the concatenation operation. 

\begin{figure}[t]
	\centering
	\subfigure[]{
		\label{fig:down}
		\includegraphics[width=0.29\textwidth]{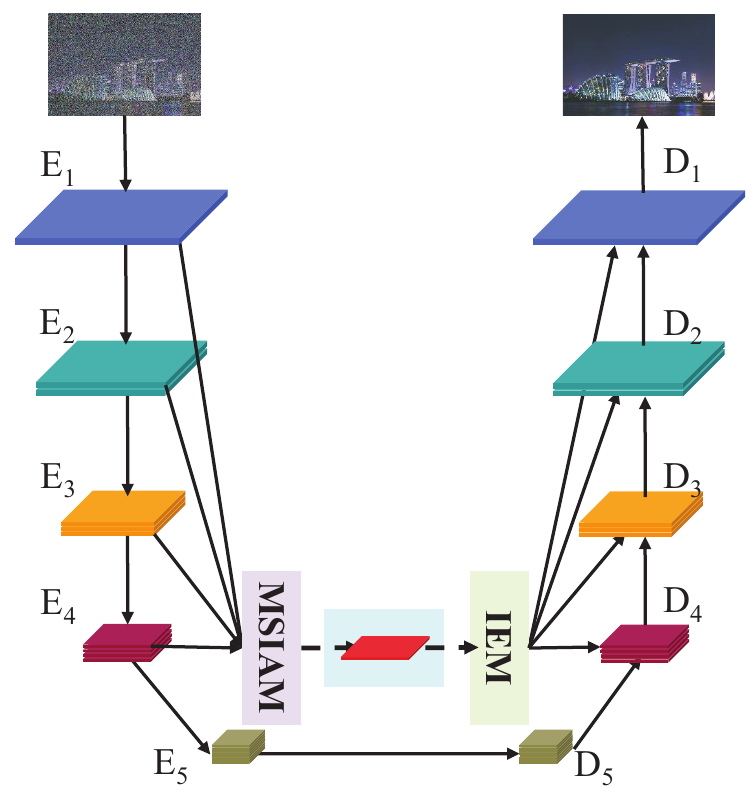}}
	\subfigure[]{
		\label{fig:mid}
		\includegraphics[width=0.29\textwidth]{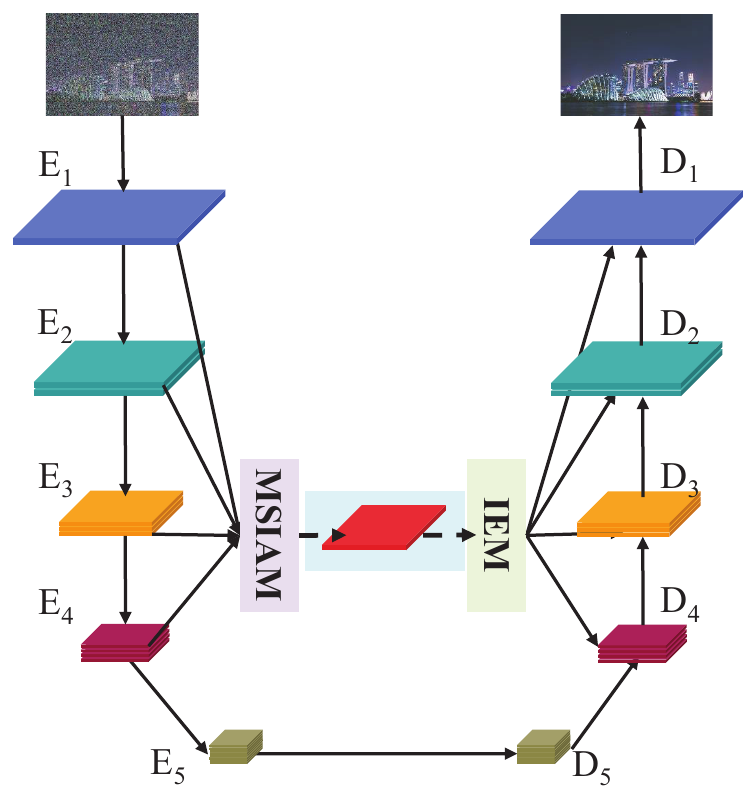}}
	\subfigure[]{
		\label{fig:up}
		\includegraphics[width=0.335\textwidth]{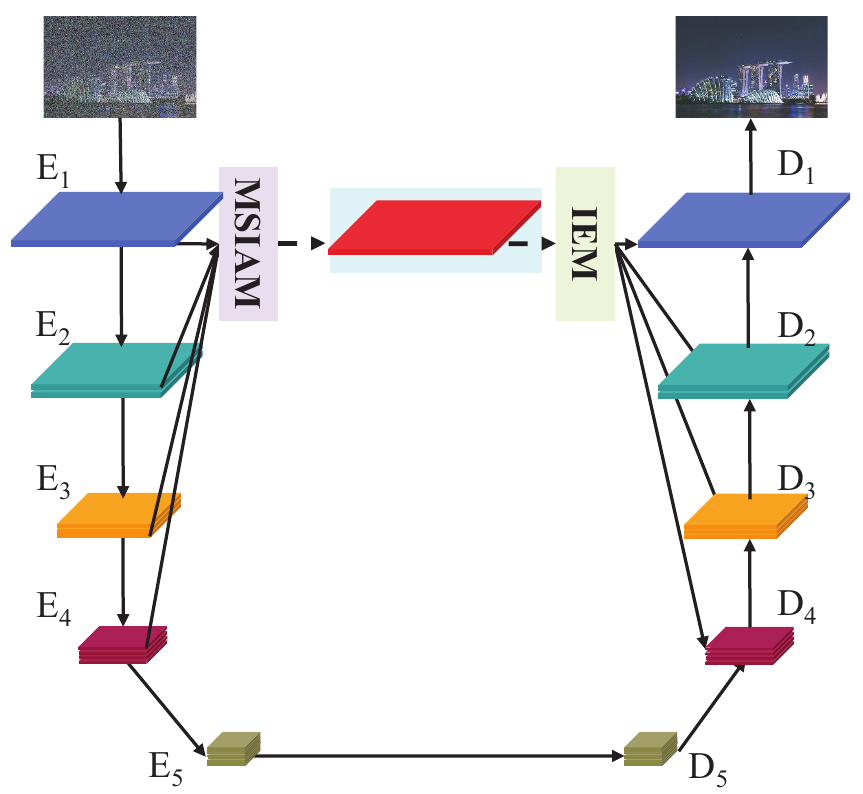}}
	\caption{Model structure variants corresponding to different target resolution. Feature maps in the blue shadow are needed to be maintained in memory. (a) The resolution of the aggregated feature maps equals to the minimum resolution in the encoder.  (b) The resolution of the aggregated feature maps equals to the intermediate resolution in the encoder. (c) The resolution of the aggregated feature map equals to the  maximum resolution in the encoder.}
	\label{fig:structure_variants}
\end{figure}

Firstly, the multi-scale feature maps is pruned to be slim. It can be realized by point-wise convolution to avoid a large computational cost. Then these multi-scale feature maps need to be resized to have the same resolution. The target resolution can be any of the multi-scale feature maps generated by the end of the encoding process. We illustrate three classical types of architecture that correspond to different target resolutions in \cref{fig:structure_variants}. For model in \cref{fig:down}, the resizing is performed by Pixel Unshuffle, for model in \cref{fig:up}, the resizing is realized by Pixel Shuffle, and for model in \cref{fig:mid}, the resizing is realized by both. The resized feature maps are further concatenated along the channel dimension. Finally, the information interaction is performed by a point-wise convolution $PWconv$. 

The parameters in the MSIAM are optimized during the training process, which are used to weight different feature maps. In this way, the information ratio between different feature maps will automatically be adjusted. Therefore, multi-scale fine-grained details are interactive and complementary to each other, leading to compact but enriched feature maps.

\subsubsection{Information Enhancement Module (IEM)} 
IEM is designed to generate enhanced multi-scale feature maps according to the compact single-scale feature maps produced by MSIAM. 

As shown in \cref{fig:overview}, IEM consists of a resize operation and an enhancement header. Firstly, the input single-scale feature map is resized to a specific resolution according to the current decoder stage. The resize operation can be an unsampling or a downsampling operator, which depends on the relative sizes of the resolutions between source and target feature maps. Accordingly, there are three classical types of IEM in \cref{fig:structure_variants}. For the network architecture in \cref{fig:overview}, the resize operation is applied with the pixel shuffle operation. 
The enhancement header consists of a ConvNeXt V2\cite{Woo2023ConvNeXtV2} block and a separable convolution. We apply the ConvNeXt V2 block due to its powerful representative ability. The separable convolution adjusts the number of output channels, which should be consistent with the channels output in the corresponding encoding stage.

To verify that the generated feature maps by IEM are not imitations of the feature maps output by the encoding process but with enhanced information, we first compute the similarity between these two groups of feature maps. We apply SSIM\cite{1284395SSIM} as the similarity metric, within the range of [0,1]. The result is shown in \cref{tab:similarity}. It shows that the feature maps generated by IEM are different from the encoder outputs. 
\setlength{\tabcolsep}{5pt}
\begin{table}[tb]		
	\centering
	\begin{minipage}[b]{0.45\linewidth}
		\centering
		\caption{The similarity relationship between the output feature maps of encoder and IEM. A larger SSIM represents greater similarity.
		}
		\begin{tabular}{@{}ll@{}}				
			\toprule
			Stage & SSIM \\
			\midrule
			\{$E_1, IEM_1$\}& 0.1813\\
			\{$E_2, IEM_2$\}& 0.0179\\
			\{$E_3, IEM_3$\}& 0.0131\\
			\{$E_4, IEM_4$\}& 0.0157\\
			\bottomrule
		\end{tabular}
		\label{tab:similarity}
	\end{minipage}
	\hfill
	\begin{minipage}[b]{0.45\linewidth}
		\centering	
		\caption{Comparison of representative ability between the output feature maps of encoder and IEM, where RA stands for the representative ability.}		
		\begin{tabular}{@{}lll@{}}				
			\toprule
			Stage & RA of Encoder & RA of IEM \\
			\midrule
			1 & 0.3315 & 0.4185\\
			2 & 0.2768 & 0.6529\\	
			3 & 0.2203 & 0.3662\\
			4 & 0.2209 & 0.2460\\											
			\bottomrule
		\end{tabular}
		\label{tab:representative_ability}
	\end{minipage}
\end{table}

To compare the representative ability of feature maps, the method used in \cite{Woo2023ConvNeXtV2} is modified. For the feature maps generated by the encoder and IEM, we calculate the cosine distance between any two channels and then calculate the mean and variance values of these distances, respectively. Based on similar mean values, the larger the variance, the greater the difference between channels, indicating a richer diversity in feature maps. The result is shown in \cref{tab:representative_ability}. We can observe that the feature maps generated by IEM is embedded with enhanced information which will boost the model's performance.

\section{Experiments and Analysis} 

\subsubsection{Details of NAFNet with UNet$--$} 
We apply the proposed UNet$--$ into NAFNet\cite{chen2022simple}, which is one of the SOTA models in the field of image restoration. NAFNet adopts a typical U-Net structure, with the encoder consisting of 5 stages \{$E_1, E_2, E_3, E_4, E_5$\}. Among them, the feature maps outputted by the first 4 stages need to be connected to the decoder via skip connections. They respectively output feature maps with channel numbers of \{$32, 64, 128, 256$\} and resolutions of \{$H*W, \frac{H}{2}*\frac{W}{2}, \frac{H}{4}*\frac{W}{4}, \frac{H}{8}*\frac{W}{8}$\}. Here, $H$ and $W$ represent the height and width of the input images.

In MSIAM, using the reduction ratio of  \{$\frac{1}{16}, \frac{1}{16}, \frac{1}{16}, \frac{1}{8}$\}, we first decrease the number of channels output by the first 4 stages to \{$2, 4, 8, 32$\}, respectively. Then we apply pixel unshuffle operators with stride = \{$8, 4, 2$\} to decrease resolutions of \{$E_1, E_2, E_3$\} to $\frac{H}{8}*\frac{W}{8}$. Finally, to obtain single-scale feature maps we concatenate them with $E_4$ along the channel dimension, and further aggregate all data through point-wise convolution. 

In IEM, according to the single-scale input feature maps generated by MSIAM, we first use pixel shuffle operators with stride = \{$8, 4, 2$\} to restore the resolution of the first 3 stages \{$E_1, E_2, E_3$\}, respectively. Then, we use four enhancement header blocks to generate multi-scale feature maps separately on the input single-scale feature maps and the newly generated feature maps with higher resolution. The final generated multi-scale feature maps share the same channel number and resolution as \{$E_1, E_2, E_3, E_4$\}. 

\subsection{Image denoising}
Noise can be introduced into an image during acquisition or processing. Image denoising aims to restore an image to its original quality by reducing or removing the noise, while preserving the important features of the image. 	

We use SIDD dataset\cite{abdelhamed2018a} as our training and test dataset, which has 30,000 noisy images from 10 scenes under different lighting conditions using five representative smartphone cameras. These images are cropped non-overlapping and generated 31888 patches whose size is 256$\times$256. These patches are divided into 30608 training patches and 1280 validation patches.

During the training process, the model is optimized with PSNR loss and Adam optimizer ($\beta_1=0.9$, $\beta_2=0.9$, weight decay$=0$) for total 200K iterations. The initial learning rate is $1e^{-3}$ which is reduced to $1e^{-7}$ with the cosine annealing schedule\cite{articleSGDR}. The training batch size is set to 32.
We use the Peak-Signal-to-Noise Ratio (PSNR) and the Structural Similarity Index  (SSIM)\cite{1284395SSIM} as metrics to evaluate our models.

\setlength{\tabcolsep}{3pt}
\begin{table}[b]
	\caption{Models and their performance in image denoising. SC is the abbreviation for skip connection. Output feature map of $E_1$: 256 × 256, 32 channels, 8bit.}
	\label{tab:denoising}
	\centering
	\begin{tabular}{@{}llllll@{}}
		\hline
		Model & PSNR & SSIM & $M_{sc}$(MB) & MACs(G) &Params(MB)\\
		\hline
		
		NAFNet\cite{chen2022simple}  & 39.97  & 0.960 & 3.75 &16.23 &29.16\\			
		NAFNet w/o SC & 39.61 & 0.957 & 0.00 &16.23 &29.16\\	
	
		\multirow{2}{2cm}{NAFNet with \\UNet$--$}& \multirow{2}{*}{40.01{\fontsize{5}{5} ($\uparrow 0.04$)}}& \multirow{2}{*}{0.960 {\fontsize{5}{5} ($\uparrow 0.0$)}}& \multirow{2}{*}{0.25(\fontsize{5}{5}$\downarrow 93.3\%$)}&
		\multirow{2}{*}{17.52(\fontsize{5}{5} $\uparrow 7.9\%$)}&
		\multirow{2}{*}{29.98(\fontsize{5}{5} $\uparrow 2.8\%$)} \\
		
		&&&&&\\		
		\hline
		KBNet\cite{Zhang2023kbnet}  & 40.35 & 0.962 & 7.5 &58.19 &104.93\\
		\multirow{2}{2cm}{KBNet with \\UNet$--$}& \multirow{2}{*}{40.39{\fontsize{5}{5} ($\uparrow 0.04$)}}& \multirow{2}{*}{0.962 {\fontsize{5}{5} ($\uparrow 0.0$)}}& \multirow{2}{*}{0.5(\fontsize{5}{5}$\downarrow 93.3\%$)}&
		\multirow{2}{*}{63.11(\fontsize{5}{5} $\uparrow 8.4\%$)}&
		\multirow{2}{*}{108.12(\fontsize{5}{5} $\uparrow 3.0\%$)} \\

		&&&&&\\													
		\hline
	\end{tabular}
\end{table}
\begin{figure}[h]
	\centering					
	\includegraphics[width=0.9\textwidth]{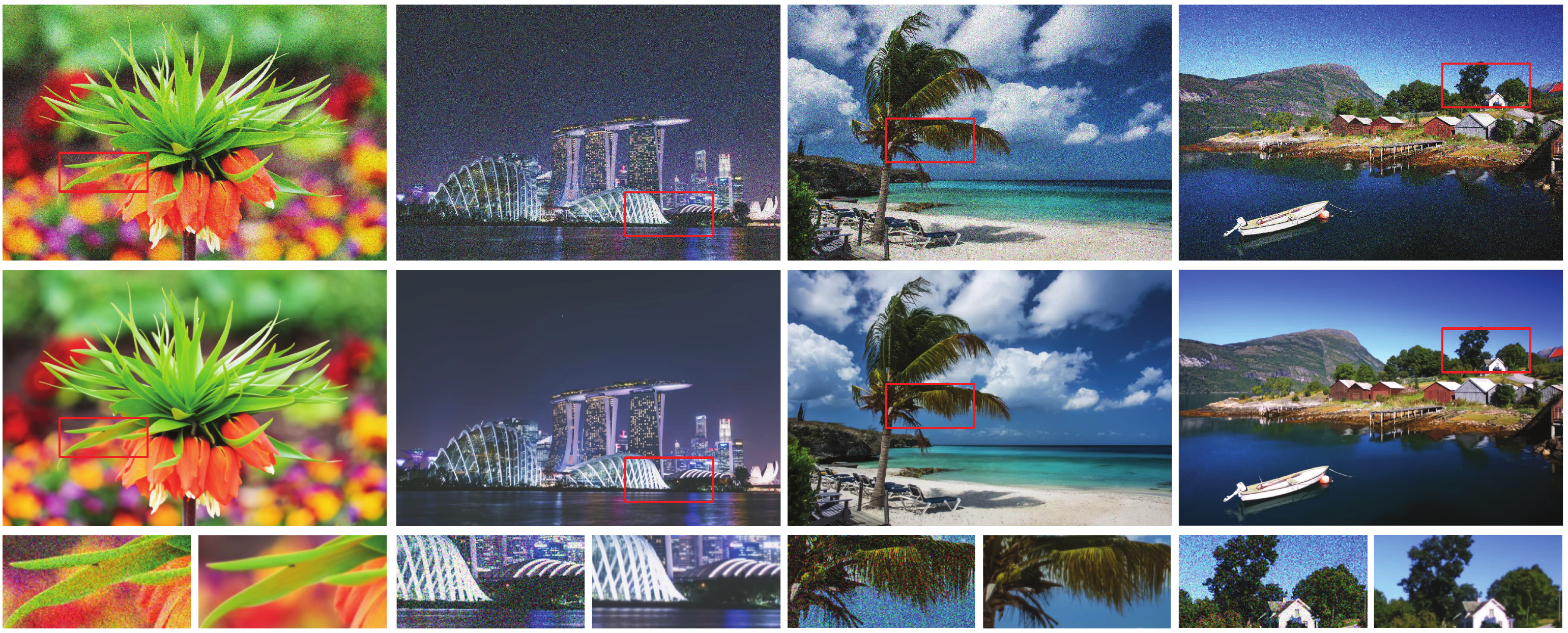}		
	\caption{Visualization results. The first row includes noisy images, the second row includes clean images outputted by NAFNet\cite{chen2022simple} with UNet$--$, the third row shows the comparison of local details. }
	\label{fig:denoising_vision}
\end{figure}

In \cref{tab:denoising}, we first compare the proposed UNet$--$ against the baseline model NAFNet and KBNet, respectively. Our model reduces the memory consumption caused by skip-connections by 93.3\%, however the performance is improved by 0.04 in PSNR in both experiments. It is worth noting that the MACs and the number of parameters increase only by 7. 9\% and 2. 8\%, 8.4\% and 3.0\%, respectively. This indicates that the proposed UNet$--$ can reduce the memory cost of skip-connections without causing harm to the model performance in an efficient way.	
We also compare the proposed NAFNet with UNet$--$ against the NAFNet without skip-connections. From \cref{tab:denoising}, we can observe that our model outperforms the model without any skip-connection by 0.4 in PSNR and 0.003 in SSIM.

\subsection{Image deblurring}
Image deblurring is one of the image processing tasks that involves removing the blurring artifacts from images or videos to restore the original sharp content. 

For image deblurring, we validate the performance of  NAFNet with UNet$--$ on GoPro\cite{8099518} dataset. It provides pairs of realistic blurry images and the corresponding sharp ground-truth images that are obtained by a high-speed camera. The blurry image is produced by averaging successive latent frames, while the sharp image is defined as the mid-frame among the sharp frames that are used to make the blurry image. There are 3,214 blurred images with the size of $1280\times720$ that are divided into 2,103 training images and 1,111 test images. 
\begin{table}[h]
	\caption{Models and their performance in image deblurring. SC is the abbreviation for skip connection. Output feature map of $E_1$: 256 × 256, 32 channels, 8bit.}
	\label{tab:deblurring}
	\centering
	\begin{tabular}{@{}llllll@{}}
		\hline
		Model & PSNR & SSIM & $M_{sc}$(MB) & MACs(G) &Params(MB)\\
		\hline

		NAFNet\cite{chen2022simple}  & 32.87  & 0.960 & 3.75 &16.23 &29.16\\
		Tailor\cite{Tailor}  & 32.87  & 0.960 & 0.00 &16.23 &29.16\\			
		NAFNet w/o SC  & 32.52 & 0.958 & 0.00 &16.23 &29.16\\
		\multirow{2}{2cm}{NAFNet with \\UNet$--$}& \multirow{2}{*}{33.06{\fontsize{5}{5} ($\uparrow 0.19$)}}& \multirow{2}{*}{0.962{\fontsize{5}{5} ($\uparrow 0.002$)}}& \multirow{2}{*}{0.25(\fontsize{5}{5}$\downarrow 93.3\%$)}&
		\multirow{2}{*}{17.52(\fontsize{5}{5} $\uparrow 7.9\%$)}&
		\multirow{2}{*}{29.98(\fontsize{5}{5} $\uparrow 2.8\%$)} \\
		
		&&&&&\\			
		\hline
	\end{tabular}
\end{table}

\begin{figure}[htbp]
	\centering					
	\includegraphics[width=0.9\textwidth]{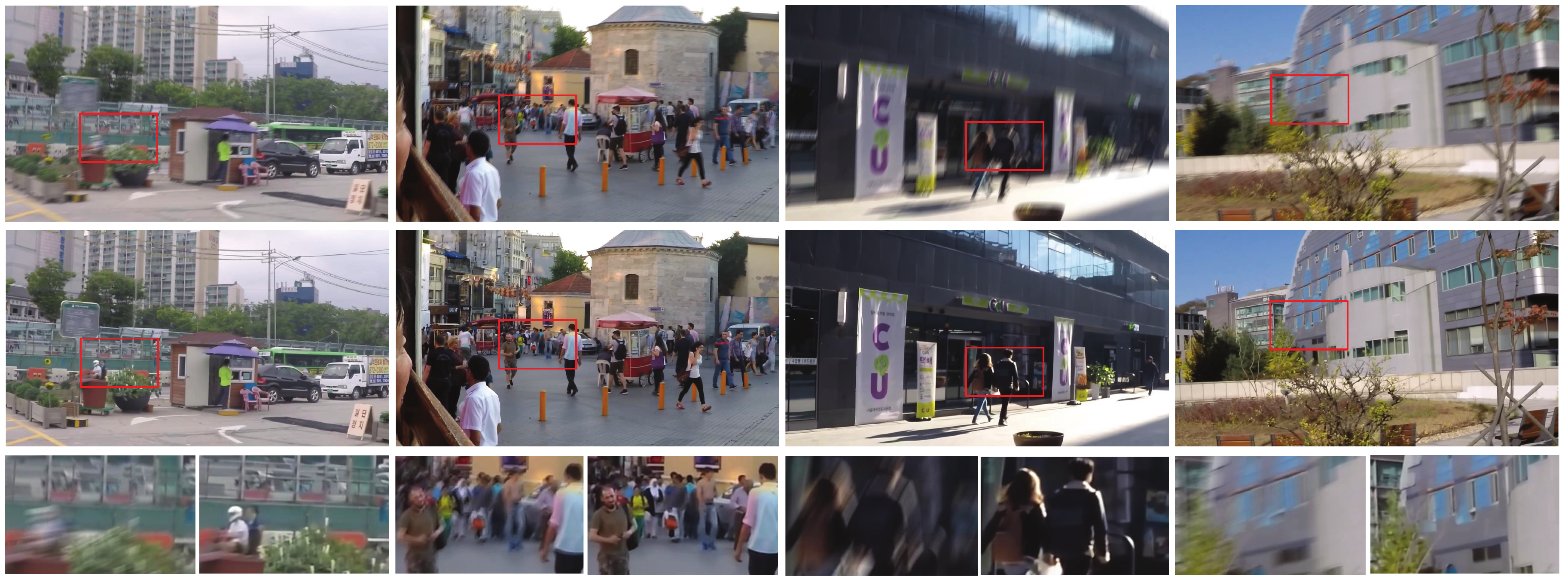}		
	\caption{Visualization results. The first row includes blurry image, the second row includes clean images outputted by NAFNet\cite{chen2022simple} with UNet$--$, the third row shows the comparison of local details. }
	\label{fig:deblurring_vision}
\end{figure}

The evaluation result is shown in \cref{tab:deblurring}. With the same $93.3\%$ memory savings, the proposed NAFNet with UNet$--$ outperforms the baseline model with 0.19 in PSNR and 0.002 in SSIM. 

\subsection{Image super-resolution}
Image super-resolution aims to increase the resolution of an image, while maintaining its content and details as much as possible. It can be used for various applications, such as improving image quality, enhancing visual detail, and increasing the accuracy of computer vision algorithms.
We train the model using the exact same network as image denoising and image deblurring. We use DIV2K\cite{DIV2K} and Flickr2K as the training dataset. We use five commonly used datasets including Set5\cite{Set5}, Set14\cite{set14}, B100\cite{BSD}, Urban100\cite{Urban100} and Manga109\cite{Manga109} as the test dataset. PSNR and SSIM are used to evaluate the quality of the restored images.
We use the AdamW optimizer with $\beta_1$= 0.9 and $\beta_2$ = 0.99 to solve the proposed model. The number of iterations is set to 500,000. We also set the initial learning rate to $1e-3$ and the minimum to $1e-5$, which is updated by the Cosine Annealing scheme.

\setlength{\tabcolsep}{2pt} 	
\begin{table}[h]
	\caption{Models and their performance in image super resolution.SC is the abbreviation for skip connection. 
	}
	\label{tab:img_resolution}
	\centering
	\begin{tabular}{p{2cm}|p{0.8cm}|p{0.85cm}|p{0.8cm}|p{0.85cm}|p{0.8cm}|p{0.85cm}|p{0.9cm}|p{0.95cm}|p{0.9cm}|p{0.95cm}}
		\hline
		\multirow{2}{*}{Model}  & \multicolumn{2}{c|}{Set5\cite{Set5}}  & \multicolumn{2}{c|}{Set14\cite{set14}}  & \multicolumn{2}{c|}{B100\cite{BSD}}  &\multicolumn{2}{c|}{Urban100\cite{Urban100}} & \multicolumn{2}{c}{Manga109\cite{Manga109}}\\ \cline{2-11}
		& PSNR&SSIM  & PSNR&SSIM  & PSNR&SSIM &PSNR&SSIM & PSNR&SSIM\\
		\hline
		
		NAFNet\cite{chen2022simple}& 37.74& 0.9603& 33.47&0.9179& 32.16& 0.8997&  31.98& 0.9272&37.96& 0.9770\\
		
		\multirow{2}{2cm}{NAFNet w/o SC} & \multirow{2}{*}{36.89}& \multirow{2}{*}{0.9588}& \multirow{2}{*}{32.33}&\multirow{2}{*}{0.9002}& \multirow{2}{*}{31.11}& \multirow{2}{*}{0.8798}&  \multirow{2}{*}{31.01}& \multirow{2}{*}{0.9132}&\multirow{2}{*}{36.81}& \multirow{2}{*}{0.9663}\\
		&&&&&&&&&&\\
		\multirow{2}{2cm}{NAFNet with \\UNet$--$}& \multirow{2}{*}{38.23}& \multirow{2}{*}{0.9611}& \multirow{2}{*}{33.96} &\multirow{2}{*}{0.9207} &\multirow{2}{*}{32.73} &\multirow{2}{*}{0.9017} &\multirow{2}{*}{32.45}  &\multirow{2}{*}{0.9321} &\multirow{2}{*}{37.93} &\multirow{2}{*}{0.9768} \\
		&&&&&&&&&&\\
		
		\hline
	\end{tabular}
\end{table}
\begin{figure}[t]
	\centering					
	\includegraphics[width=0.88\textwidth]{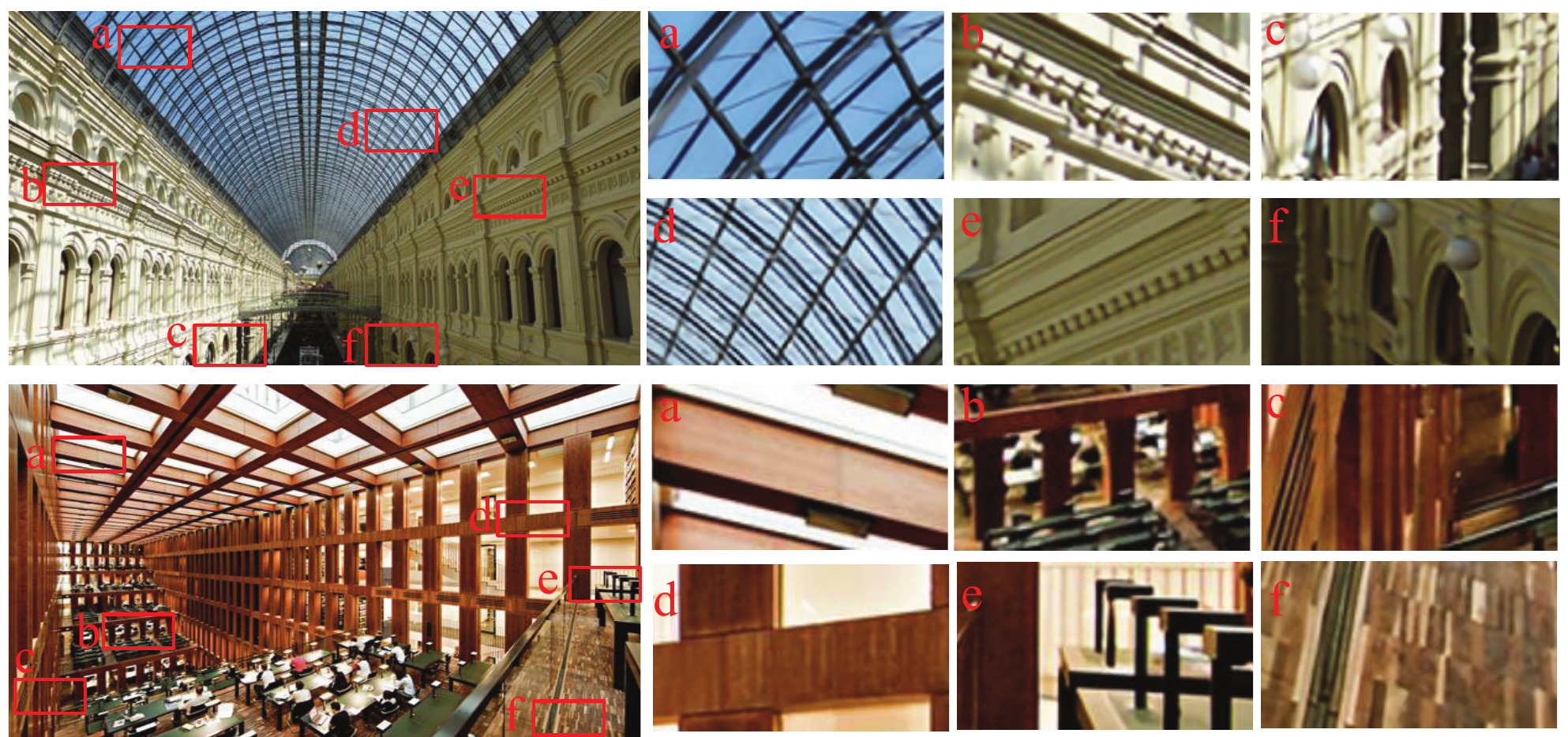}		
	\caption{Visualization results of the proposed NAFNet\cite{chen2022simple} with UNet$--$ on image super resolution.}
	\label{fig:SR_vision}
\end{figure}

For each model, we calculate the average values of PSNR and SSIM on five datasets. With the same $93.3\%$ memory savings, the proposed NAFNet with UNet$--$ outperforms the baseline model with 0.398 in PSNR and 0.002 in SSIM.

\subsection{Image matting}
Image matting is a fundamental computer vision task which aims to predict an alpha matte to precisely cut out an image region. It estimates the foreground, background, and alpha matte from a single image, but has many applications in image and video editing, virtual reality, augmented reality, entertainment.

Unlike above image restoration tasks, for image matting task we use $\rm MSCAN_{tiny}$ \cite{Guo2022SegNeXtRC} as a new backbone of the encoder.

We apply our proposed method to obtain $\rm MSCAN_{tiny}$ with UNet$--$. We use P3M-10K\cite{P3M} as the training dataset. It consists of about 10,000 high resolution face-blurred portrait images which are carefully collected and filtered from a huge number of images with diverse foregrounds, backgrounds and postures. In addition, face obfuscation is used as a privacy protection technique to remove identifiable face information while retaining fine details such as hairs. We use three commonly used datasets including P3M-500-P\cite{P3M}, P3M-500-NP\cite{P3M} and RealWorldPortrait-636\cite{Mask_Mask_matting} as the test dataset. 

We use SAD and MSE metrics to evaluate models on the whole image and the unknown region defined by trimap images. SAD and MSE refer to the sum of absolute differences and the mean squared error between the ground-truth alpha matte and the predicted alpha matte, respectively. 

\begin{table}[t]
	\caption{Models and their performance in image matting. SC is the abbreviation for skip connection. The unit represented by $^*$ is $\rm1e-3$. 
	}
	\label{tab:img_matting}
	\centering
	\begin{tabular}{@{}c|c|c|c|c|c|c|c|c@{}}
		\hline
		\multirow{2}{*}{Model} & \multirow{2}{0.8cm}{$M_{sc}$\\(MB)} & \multirow{2}{0.9cm}{MACs\\(G)} &\multirow{2}{1.1cm}{Params\\(MB)} & \multirow{2}{*}{Dataset}  &\multicolumn{2}{c|}{whole image}  &\multicolumn{2}{c}{Unknown}  \\ \cline{6-9}
		&&&&& MSE$^*$ & SAD& MSE$^*$ & SAD\\
		\hline
		
		\multirow{3}{*}{$\rm MSCAN_{tiny}$\cite{Guo2022SegNeXtRC}} & \multirow{3}{*}{2.56}& \multirow{3}{*}{1.57}& \multirow{3}{*}{3.95}&RWP-636\cite{Mask_Mask_matting} &14.107&7.519&63.723&5.431\\ \cline{5-9}
		&&&& P3M-500-NP\cite{P3M} & 2.369&1.619&14.926&1.296\\ \cline{5-9}
		&&&& P3M-500-P\cite{P3M} &2.998&1.736&18.735&1.295\\ 
		\hline

		\multirow{3}{2cm}{$\rm MSCAN_{tiny}$ w/o SC}  & \multirow{3}{*}{0.00}  & \multirow{3}{*}{1.55}  & \multirow{3}{*}{3.95} &RWP-636\cite{Mask_Mask_matting} &12.423&5.925&60.345&3.112\\\cline{5-9}			
		&&&& P3M-500-NP\cite{P3M} &1.645&1.321&13.994&1.076\\\cline{5-9}
		&&&& P3M-500-P\cite{P3M} &2.386&1.184&16.842&1.130\\
		
		\hline
		\multirow{3}{2cm}{$\rm MSCAN_{tiny}$ with UNet$--$} & \multirow{3}{*}{0.14}  & \multirow{3}{*}{1.71}  & \multirow{3}{*}{4.23} &RWP-636\cite{Mask_Mask_matting} &14.835&7.611&64.192&5.587\\\cline{5-9}
		
		&&&& P3M-500-NP\cite{P3M} &2.391&1.689&15.164&1.325\\\cline{5-9}
		
		&&&& P3M-500-P\cite{P3M} &3.042&1.728&19.236&1.313\\
		\hline

	\end{tabular}
\end{table}
\begin{figure}[t]
	\centering					
	\includegraphics[width=\textwidth]{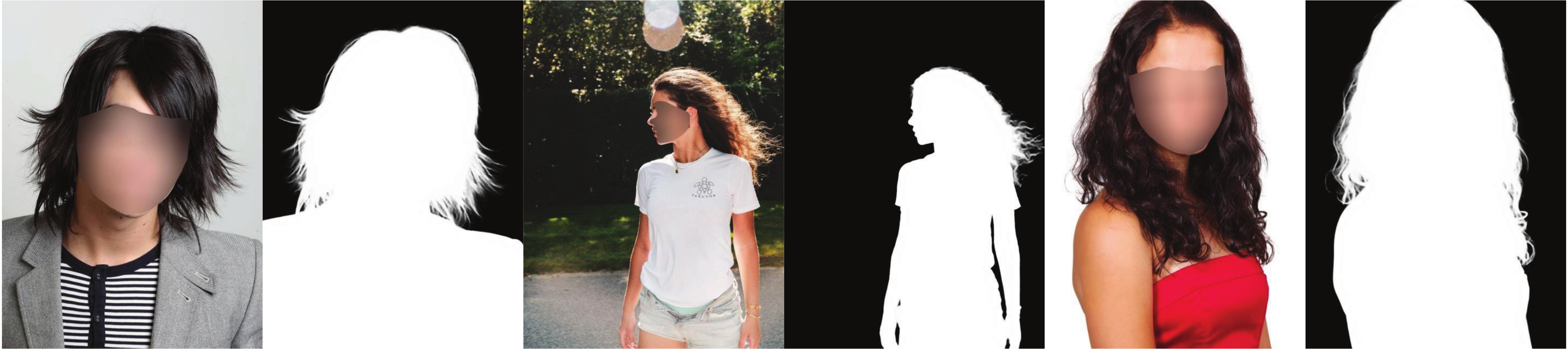}		
	\caption{Visualization results of the proposed $\rm MSCAN_{tiny}$\cite{Guo2022SegNeXtRC} with UNet$--$ 
		on image matting.}
	\label{fig:matting_vision}
\end{figure}
For each model, we calculate the average values of MSE and SAD on three datasets. With the $94.5\%$ memory savings, the proposed $\rm MSCAN_{tiny}$ with UNet$--$ outperforms the baseline model $\rm MSCAN_{tiny}$ with 0.246 (whole image), 0.402 (unknown region) in MSE and 0.051 (whole image), 0.067 (unknown region) in SSIM, respectively. Some visualization results are shown in \cref{fig:matting_vision}.

\section{Conclusion}
We have presented an effective method to reduce the memory demand for skip-connections in U-Net and its variants. In addition, we have obtained more enriched information based on the feature map transmitted by skip-connection. Applying the method, we construct a memory-efficient and feature-enhanced network architecture, UNet$--$. It is shown that the approach achieves better or comparable results compared to the models with common skip-connections on various kinds datasets for different applications, including image denoising, image deblurring, image super resolution, and image matting. 

\newpage
 \bibliographystyle{splncs04}
 \bibliography{0633}

\end{document}